# DropRegion: Training of Inception Font Network for High-Performance Chinese Font Recognition

**Shuangping Huang**  huangshuangping@gmail.com
**Zhuoyao Zhong**  z.zhuoyao@mail.scut.edu.cn
**Lianwen Jin**  lianwen.jin@gmail.com
**Shuye Zhang**  shuye.cheung@gmail.com
**Haobin Wang**  klvn930815@gmail.com
School of Electronic and Information Engineering, South China University of Technology, Guangzhou 510641, China

**Abstract**

Chinese font recognition (CFR) has gained significant attention in recent years. However, due to the sparsity of labeled font samples and the structural complexity of Chinese characters, CFR is still a challenging task. In this paper, a DropRegion method is proposed to generate a large number of stochastic variant font samples whose local regions are selectively disrupted and an inception font network (IFN) with two additional convolutional neural network (CNN) structure elements, i.e., a cascaded cross-channel parametric pooling (CCCP) and global average pooling, is designed. Because the distribution of strokes in a font image is non-stationary, an elastic meshing technique that adaptively constructs a set of local regions with equalized information is developed. Thus, DropRegion is seamlessly embedded in the IFN, which enables end-to-end training; the proposed DropRegion -IFN can be used for high performance CFR. Experimental results have confirmed the effectiveness of our new approach for CFR.

## 1 Introduction

Font recognition plays an important role in document analysis and recognition of character images. It is also fundamental to graphic design, as font is a core design element of any printed or displayed text. However, font recognition has long been neglected by the vision community, despite its importance. Font recognition is an inherently difficult and error-prone task because of the huge number of available fonts, the dynamic and open-ended properties of font classes, and the very subtle and character-dependent differences among fonts [1]. Over the years, some approaches to font recognition have been proposed [2-8], but most font recognition research has been carried out on text in Western languages, and the approaches proposed have not yielded satisfactory accuracy. Because of the structure and ideographic nature of individual Chinese characters, Chinese font recognition is more challenging than recognition of most Western scripts. The limited number of previous studies [5, 9-10, 12, 32] of Chinese font recognition have mostly approached the problem from a document analysis perspective, which makes the results highly sensitive to noise and only applicable to simple cases with strong constraints. In this study, therefore, we approached the development of a more effective automatic Chinese font recognition method from a computer vision perspective.

Recently, deep convolutional neural networks (CNNs) have achieved great success in many computer vision tasks, such as image classification [13-17], handwritten character recognition [18-21, 25], and object detection [22-23, 26-27]. These CNN methods receive raw data and automatically learn the representations needed for specific tasks. Beyond the conventional pattern analysis pipeline, which involves feature extraction and a classifier design, a CNN is a type of end-to-end representation learning framework. Font recognition, on the other hand, can be regarded as a special form of image classification problem. Based on this analysis, we propose that Chinese font recognition is a representation learning problem that can be modeled by CNNs. The current explosion of research on the use of CNNs for deep learning in computer vision has produced many good examples of CNN networks, including AlexNet [13], the Zeiler & Fergus model [14], VGGNet [15], GoogleNet [16] and others. However, use of these networks is not guaranteed to result in improved performance. Careful design of a CNN-based network is required for the specific task of Chinese font recognition.

An inception-type network is at the core of most state-of-the-art computer vision solutions based on CNNs [16, 18-19]. This inception architecture has made it feasible to use CNN-based solutions in big-data scenarios to achieve higher performance without adding extra complexity. The basic inception-style building block introduces parallel structures that run convolutional layers on several different scales, plus an extra pooling layer, and concatenates their responses. In this way, the impact of structural changes on nearby components is mitigated to adapt to the diversity of data samples. Chinese characters are characterized by their elaborate structures. These structures are rendered on different scales of local patches, comprehensively reflecting font identification information. Therefore, it is reasonable to assume that inception is a suitable network structure choice for Chinese font recognition. The general design principles of inception-type networks have been presented in previous studies [16, 18]. Research to date has not shown, however, that use of an inception-type network, e.g., GoogleNet, will yield significant quality gains in practical scenarios. To this end, an inception font network (IFN) was constructed in this study specifically for use in Chinese font recognition.

It has been determined empirically that a classifier can distinguish characters correctly even when they are partially blocked. This capability is similar to that of human beings, who can correctly discriminate a visual object using the surrounding context. In other words, regardless of whether a local region exists, the font category is not changed. Inspired by the above observation and analysis, we sought to improve generalization performance by introducing disrupted samples during the training phase of the IFN. In this study, we formulated this occlusion process as DropRegion. Specifically, during the training process, one image is divided into several regions. Then, one or more local regions are randomly selected and destructed by a type of noise. This process is embedded in the IFN, and the whole framework can be trained end to end by back propagation and stochastic gradient descent. It has been observed that printed characters pose a variety of spatial layouts, while the stroke information over the character image centroid is significantly more compact than at the boundary. To cater to the distribution characteristics of character structures, we developed an elastic mesh technique for programming the region division in DropRegion. Basically, an elastic mesh technique is used to divide an image elastically into several parts, ensuring that each part contains an equal amount of information.

To summarize, we present a new approach to Chinese font recognition, DropRegion-IFN, that integrates an IFN and a new model regularization method called DropRegion, which randomly removes several elastically sized regions from the characters of an original Chinese character prototype while retaining the identity information contained in it. The proposed DropRegion implements data augmentation, thus improving the generalized applicability of the CNN-based network model and preventing model overfitting. Furthermore, we introduce IFN for sufficient feature representation, which caters to the DropRegion training mode and specifically considers the Chinese font recognition task.

The remainder of this paper is organized as follows. Related research is summarized in Section 2. The proposed DropRegion-IFN method is described in Section 3. Single character-based and text block-based font recognition schemes are described in Sections 4 and 5, respectively. Experimental results are presented in Section 6. Conclusions drawn from the results of the research are presented in Section 7.

## 2 Related studies

Researchers typically regard font recognition as a pattern recognition task that involves feature extraction and classifier design. Various methods that emphasize feature extraction or classifier design have been proposed in the font recognition field [6-10]. For example, Cooperman [6] used local detectors to identify regions that have typographic properties (e.g., serif, boldness, and italics). Zramdini et al. [7] employed a scale-invariant feature transform algorithm to build an Arabic font recognition system. Zhu et al. [8] extracted the texture features of a text block containing several characters using a group of Gabor filters. Ding et al. [9] extracted wavelet features from a text image using a method that yielded good performance with text images of different sizes. Tao et al. [10] used multiple local binary patterns to describe the discriminative information in a text block. These studies focused on extraction of different features, which suggests that typeface feature descriptors dictate the accuracy of font recognition. However, useful information can easily be lost because of the fixed handcrafted feature representation rule. Several attempts to improve overall classification performance have involved designing complex learning algorithms that are performed on extracted features. Zhu et al. [8] applied a weighted Euclidean distance classifier to Chinese font information. Ding et al. [9] used modified quadratic discriminant functions to enhance classification power. Slimane et al. [11] used Gaussian mixture models to build an Arabic font recognition system. Zhang et al. [12] used a support vector machine to distinguish among 25 types of Chinese fonts. All of these studies involved the usual approach of feature extraction followed by classifier design, which requires careful engineering and considerable domain expertise and separates the two stages completely.

Going beyond the "feature extraction plus classifier design" pipeline, CNNs have been used in end-to-end approaches to jointly learn features representation and as a classifier. CNNs have been used with great success in the field of computer vision [13-16, 18-22]. However, to the best of our knowledge, there is no Chinese font recognition method based on CNNs, although one related study [1] proposed the use of CNNs for Roman alphabets. Wang et al. [1] developed the DeepFont system, which relies on hierarchical deep CNNs for domain adoption to compensate for real-to-synthetic domain gaps and improve Roman character font recognition. This system employs a basic CNN that is similar to the popular AlexNet structure for ImageNet [13]. Another system worth mentioning is the principal component 2DLSTM (2-D long short-term memory) algorithm proposed by Tao et al. [42], in which a principal component layer convolution operation is introduced to handle noisy data and take advantage of 2DLSTM to capture the contrast between a character trajectory and the background. However, this algorithm applies only to Chinese font recognition based on single characters; it is not applicable to font identification based on Chinese text blocks.

CNNs are hierarchical architectures that are stacked with multiple non-linear modules and have powerful abilities in mapping inputs to outputs. However, being neural networks, CNNs can easily become trapped in local minima when inappropriately handled. When this happens, degradation of representation learning [28] performance may occur. To improve the generalization performance or reduce overfitting, researchers have proposed several effective methods, including Dropout [29], DropConnect [30], model ensemble learning [24], and others. During the training process, Dropout randomly drops a subset of hidden neurons, while DropConnect stochastically excises a subset of connections between two hidden layers. The ensemble method is also an effective technique for reducing generalization errors by combining multiple models. Huang et al. [31] recently developed a training algorithm that drops a random subset of layers into a deep residual network [20] and achieves good performance in reducing overfitting. Bastien et al. [32] developed a powerful generator of stochastic variations for handwritten character images. Simard et al. [33] expanded the training set by adding elastically distorted samples for use in visual document analysis. In this paper, we propose a new method called DropRegion to improve generalization performance by introducing disrupted samples into the training process rather than by regularizing activations, weights, or layers. In DropRegion, some small regions are randomly dropped from the characters, and the remaining regions are recombined to form a new character. Using DropRegion, a large number of new and diverse characters can be generated from a prototype character, thereby solving the problem of scarcity of labeled font image samples. It should be noted that Gidaris et al. [35] applied multiple regions, including the original candidate box, half boxes, central regions, border regions, and contextual regions, for learning a detection-oriented model. They claimed that half boxes, which utilize each half of an object, make the features more robust to occlusions. In our approach, instead of blocking fixed parts (left, right, top, bottom) of an object, a more flexible strategy is adopted that randomly disrupts one or more small regions in each text image during the training process.

Since Google introduced the "inception module" for state-of-the-art image classification in 2014 [19], inception has been a central component of the deep convolutional neural architecture that has driven CNN development. The inception architecture has since been refined in various ways, first by the introduction of batch normalization [36] and later by additional factorization ideas [16]. Furthermore, inception architecture was combined with residual connections to develop Inception-ResNet, which won the first place in the 2016 ILSVRC classification task competition [18]. A few general principles and optimization ideas related to inception are described in these studies, but these principles do not necessarily yield performance gains in practical use. It is also impossible to apply the inception network structure proposed in the above-mentioned studies to a particular application case. To this end, we explored the potential advantages of designing an inception-integrated network for Chinese fonts, with the goal of improving CFR performance based on high-quality, learned visual features.

Cascaded cross-channel parametric pooling (CCCP) was introduced by Lin et al. [37] to allow complex and learnable interactions of cross-channel information. CCCP has considerable capabilities in modeling various distributions of latent concepts, using a network-in-network (NIN) structure to achieve better local abstraction. Another important idea proposed by Lin et al. [37] is global average pooling over feature maps, which functions as the counterpart of CCCP. Global average pooling enforces correspondence between feature maps and categories, providing a structural regularizer that reduces the effect of overfitting. In this study, we developed an IFN, equipped with inception, CCCP, and global average pooling network structure components, for use in Chinese font recognition. This approach was motivated by the following three aspects of the problem: 1) Chinese language has a large set of characters, and these characters vary widely in structure and grey-level distribution. Because Chinese characters carry font discrimination information in local structures of different scales with different character construction, inception is suitable for application to CFR because inception employs different kernel sizes to capture correlation structures of different scales. 2) Diverse and overlapping strokes can be abstracted well using CCCP. 3) The proposed DropRegion yields a large number of diverse samples to compensate for the scarcity of labeled training data. To reduce overfitting caused by network complexity, global averaging pooling is employed as a counterpart to CCCP.

## 3 Proposed DropRegion-IFN method

### 3.1 Inception font network

We have designed an IFN specifically for Chinese font recognition by combining an inception module, CCCP layers, and global average pooling. The structure of the IFN is illustrated in Fig. 1.

As Fig. 1 shows, we modified the most common type of inception module by adding three branches: one $3 \times 3$ convolution followed by two stacked $2 \times 2$ convolutions, two stacked $2 \times 2$ convolutions, and two stacked $3 \times 3$ convolutions. The first and second branches have an effective receptive field of $5 \times 5$, and the third branch has an effective receptive field of $3 \times 3$. Factorizing a larger convolution into several smaller convolutions, e.g., factorizing $5 \times 5$ into two stacked $3 \times 3$ convolutions or factoring $3 \times 3$ into two stacked $2 \times 2$ convolutions, makes it possible to increase the depth and width of the network while maintaining a consistent computational budget. This application of multi-scale convolutional kernels ($2 \times 2$, $3 \times 3$, and $5 \times 5$ convolutions) reflects the basic concept of the inception approach, i.e., that visual information should be processed at various scales and then aggregated so that the subsequent stage can abstract features simultaneously from different scales. After the inception module operation, several feature maps with the same sizes are obtained by means of a padding strategy and precise designs. Finally, the feature maps are concatenated using a concatenation layer. Because of these additional branches, more non-linear rectification layers can be incorporated into our model, enhancing the advantages and feasibility of the inception module. Furthermore, because of the inception module, we can extract local feature representations using more flexible convolutional kernels, which are filters of various sizes organized in a layer-by-layer form.

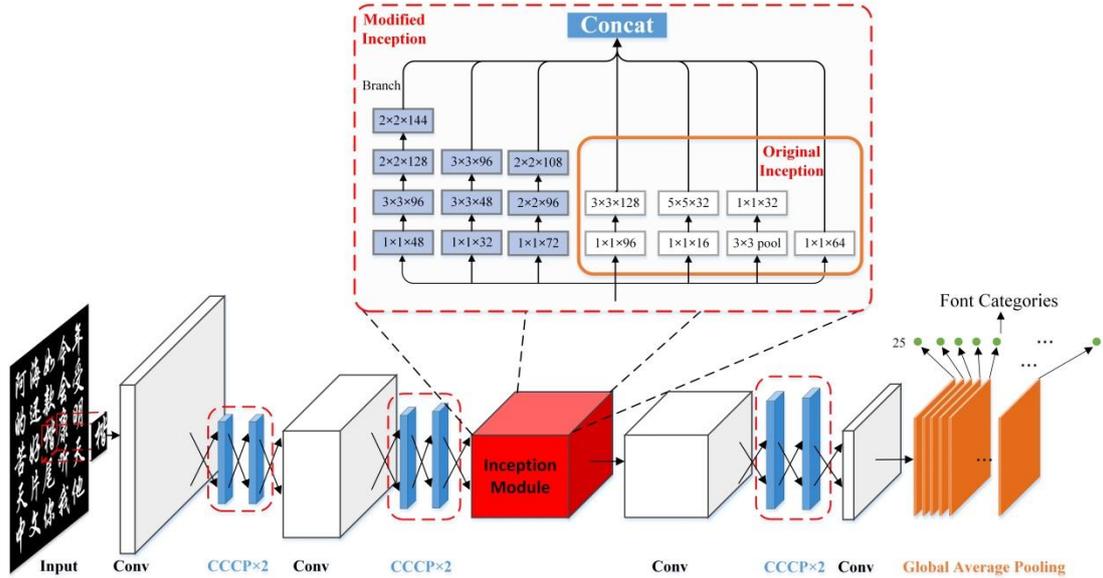

Fig. 1 Schematic illustration of IFN structure

Given that Chinese character structure is of critical importance in font recognition, we use multiple CCCP layers, as shown in Fig. 1. Pairs of adjacent CCCP layers are stacked together, and an extra traditional convolution layer is added below each pair of CCCP layers, resulting in a micro neural network (referred to as mlpconv in paper [37]) to abstract the data within the underlying patch. Nonlinear activation functions (ReLU) are embedded between the CCCP layers, increasing the level of abstraction. This micro neural network takes the place of a traditional convolution kernel, sliding over the input, obtaining the feature maps, and feeding into the next layer. This sliding mode makes the micro network shared among all the local receptive fields. As Fig. 1 shows, a total of three micro neural networks are used in our IFN, two of which are stacked below the inception module and the third of which is stacked on top of it. The layout of the micro neural network in relation to the core inception module is intended to achieve better abstraction for features of all levels. The last micro neural network is covered by an additional layer of $1 \times 1$ convolutions to reduce the representational dimension to the number of font categories (25 in this study). Thus far in the process, explicit confidence maps for all of the Chinese font categories are obtained. The final step is global average pooling for each map to obtain the resulting confidence vector that corresponds to the font categories, saving a large number of parameters. As described in [37], global average pooling can be viewed as a structural regularizer that reduces overfitting. The archi-tectural parameters tuned specifically for single character-based and text block-based Chinese font recognition are shown in Tables 1 and 2, respectively, in Section 5.

### 3.2 DropRegion

The idea of multiple regions—including the original candidate box, half boxes, central regions, border regions, and contextual regions—was used by Gidaris et al. [35] to train a detection-oriented model. They proved that using half boxes is an effective way to improve occlusion detection performance. In our approach, instead of blocking fixed parts (left/right/top/bottom) of an object, we adopt a more flexible scheme to randomly disrupt one or more small regions of a text image. First, an elastic meshing technique is used to divide a character or text block image into a set of small regions. Each region contains an equal amount of information. Next, a well-designed strategy guides the algorithm in disrupting the selected region. During the training phase, it is uncertain which regions are blocked because the algorithm permits blocking of any text image region in each iteration. Consequently, the training set size is effectively increased, and the diversity of the training samples is enriched, which reduces overfitting. More importantly, the machine is guided in learning a greater number of invariant features, regardless of the existence of a certain local part.

#### 3.2.1 Elastic meshing

Let us consider a gray-level image as a mapping, $I: \mathcal{D} \subset \mathbb{Z}^2 \to S$, where typically $S = \{0,1,\ldots,255\}$. A naive method is used to divide the image into L×L equal rectangular regions; i.e., the area of each region is equal. We call this method a fixed meshing division. However, we determined experimentally that, in most cases, the randomly selected region is the background, which does not contain character information. This is because the centroid of a Chinese character image typically contains many strokes, whereas the boundary contains few strokes. To visualize this property, 3,866 commonly used Chinese characters in 25 fonts were averaged. The final average image is shown in Fig. 2. Few strokes exist at the boundary, whereas many strokes exist at the centroid. It is interesting to note that the average character is not a Gaussian-like distribution, which was unexpected.

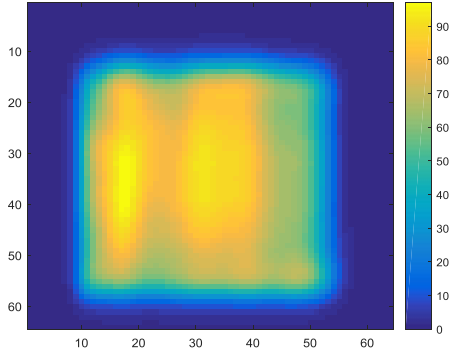

Fig. 2 Heat map of average of Chinese characters. Each character image was first inverted so that the background intensity was 0 and the stroke intensity was 255. The character images were then resized to $64 \times 64$. Finally, the element-wise average of these resized images was computed.

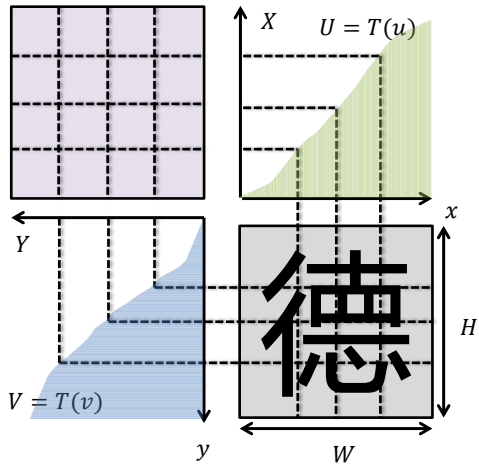

Fig. 3 Elastic meshing technique applied to one Chinese character.

To address the uneven distribution of strokes or information in Chinese characters, an elastic meshing division method was developed. This elastic meshing technique has previously been applied to shape normalization in handwritten character recognition [38-40] and has been found to be a simple and effective preprocessing technique. Elastic meshing division for shape normalization makes it possible for each mesh to contain an equal distribution of the stroke histogram. In our system, elastic meshing is used to generate a set of regions adaptively, so that the regions contain equal the geometric area, and it can be generalized to any type of feature. In this study, for simplicity, the information in each region was measured by the accumulated pixel intensity. To equalize the accumulated pixel intensities in the regions, we first calculate the projection profile histogram on each axis. Figure 3 illustrates the application of the elastic meshing technique to one character. Because the operation on the x axis is the same as that on the y axis, only the operation on the x axis is explained here. Let $\phi(x)$ denote the function for computing the projection profile histogram along the x axis, which is written as follows:

$$\phi(x) = \sum_{y=1}^{H} I(x, y), \quad (1)$$

where $H$ is the image height. Next, the image is horizontally divided into $L$ bars. We intend for the information in these bars to be equal. That is, each bar contains $\frac{1}{L}\sum_{x=1}^{W}\phi(x)$ information. Hence, the sum of the information from the first bar to the $i^{th}$ bar can be expressed as follows:

$$T(u_i) = \frac{i}{L}\sum_{x=1}^{W}\phi(x), \quad (2)$$

where $i \in \{1,..,L\}$ and $u_i$ denotes the position along the $x$ axis that corresponds to the $i^{th}$ bar. (Note that $\{u_1, u_2, ..., u_L\}$ is a set of breaking points that divides a character image into $L$ bars.) $T(u_i)$ can be formulated as an accumulated distributed function as follows:

$$T(u_i) = \sum_{x=1}^{u_i}\sum_{y=1}^{H}I(x,y). \quad (3)$$

Finally, $u_i$ is solved by substituting Eq. (2) with Eq. (1) and then substituting Eq. (3) with Eq. (2). In the same manner as $\{u_1, u_2, ..., u_L\}$ is determined, a set of breaking points along the $y$ axis, namely, $\{v_1, v_2, ..., v_L\}$, can be determined.

### 3.2.2 DropRegion strategy

Inspired by Dropout [29], we designed a similar strategy that we called DropRegion. First, an image is selected. If a random number is greater than DropRatio, $\gamma$, the image requires further processing. Second, elastic meshing is performed on the image. Third, $n$ ($n \geq 1$) local regions are randomly selected and disrupted by one type of noise. For simplicity, the randomly selected region is disrupted by multiplication using an all-zero mask. With these additional samples, the IFN can still be optimized in an end-to-end manner by stochastic gradient descent and back propagation. The proposed DropRegion algorithm is presented as Algorithm 1 at the end of this section.

To illustrate the proposed DropRegion algorithm, we consider a CNN with an input image $\mathbf{x}$, where $x \in R^{d \times d}$. When DropRegion is applied, it can be written as $M \star x$, where $\star$ is the element-wise product and M is a binary matrix that encodes the disruption information and can be expressed as follows:

$$\mathbf{M} = \begin{bmatrix} \mathbf{m}_{11} & \cdots & \mathbf{m}_{1L} \\ \vdots & \ddots & \vdots \\ \mathbf{m}_{L1} & \cdots & \mathbf{m}_{LL} \end{bmatrix}. \quad (4)$$

where $\mathbf{m}_{ij}$ is a sub-block matrix of $\mathbf{M}$. Note that some sub-block matrixes are randomly set as all-zero matrixes, whereas others are all-one matrixes. In other words, $M$ has $C_{L \times L}^n$ types of patterns and $M \sim U(0,1)$. From that starting point, the DropRegion processes an input image **x** in a convolutional manner as follows:

$$x_j^1 = g((M * x) * k_j^1 + b^1) \quad (5)$$

where $x_j^1$ is the $j^{th}$ feature map of the first convolutional layer, $k_j^1$ is the convolutional kernel connected to the $j^{th}$ feature map, $b^1$ is the bias vector of the first convolutional layer, and $g(\cdot)$ is a nonlinear activation function. The receptive field corresponding to each unit of the feature map is a region in the input image. Let **r** denote this region. If **r** is disrupted by an all-zero mask, $m$, and $b^1$ is set as an all-zero vector, then $g\left((m \star r) * k_j^1 + b^1\right) = 0$. This is because many commonly used activation functions, such as the tanh, centered sigmoid, and rectified linear units (ReLU) [41] functions, have as a property $g(0) = 0$.

The above process is illustrated in Fig. 4. The $2 \times 2$ black region in each feature map of the *Conv1* layer is obtained by convolution with 32 kernels $3 \times 3$ in size. The region is then passed to an activation function. Similarly, the $1 \times 1$ black pixels in the feature maps of the *Conv2* layer are obtained by convolution with 64 kernels $2 \times 2$ in size. The pixels are then passed to an activation function. Note that the receptive field of the black units in the *Conv2* layer is the disrupted region in the input image. That is, DropRegion causes the neurons in the deeper layers to become zeroes. Accordingly, DropRegion can be regarded as a special case of Dropout; it differs, however, in that it acts on convolutional or pooling layers.

Given input data $x$, the overall model, $f(x; \theta, M)$, outputs a result, $y$, via a sequence of operations (e.g., convolution, nonlinear activation). The final value of $y$ is obtained by summing over all possible masks, as follows:

$$y = \gamma f(x; \theta) + (1 - \gamma) \mathbb{E}_M[f(x; \theta, M)], \quad (6)$$

where $\mathbb{E}_M[f(x; \theta, M)]$ can be written as follows:

$$\mathbb{E}_M[f(x; \theta, M)] = \sum_M p(M) f(x; \theta, M). \quad (7)$$

Equation (6) reveals the mixture model interpretation of DropRegion, where the output is a combination of $f(x; \theta)$ and $f(x; \theta, M)$.

A connection exists between DropRegion and some previous data augmentation techniques. Krizhevsky et al. [13] enlarged a training set by performing tenfold cropping in one image and jittering the intensity of each image pixel. Bastien et al. [32] increased the richness of a training set using stochastic variations and noise processes (e.g., affine transformations, slant, etc.). Simard [33] expanded the training set for a neural network using an elastic transformation method. All of these methods have one characteristic in common, namely, they increase the training sample diversity and thus effectively reduce overfitting.

---

**Algorithm 1** Optimization for the DropRegion -embedded IFN

**Require**: Iteration number $t = 0$. Training dataset $\{(x_1, y_1), \ldots, (x_N, y_N)\}$, $x_n \in \mathbb{R}^D$ and $y_n \in \{0,1,\ldots,K\}$. Learning rate is $\alpha_t$.

**Ensure**:

Network parameters $\Theta$.

1: **repeat**
2:   $t \leftarrow t + 1$;
3:   Randomly select a subset of samples from training set, namely a mini-batch.
4:   **for** each training sample **do**
5:     **if** rand(0,1) < $\gamma$
6:       Perform elastic meshing division and obtain $L \times L$ regions;
7:       Randomly select a number of regions and disrupt these regions;
8:     **end**
9:     Perform forward propagation to obtain $\phi_n = f(\Theta, x_n)$.
10:   **end for**
11:   $\Delta W = 0$.
12:   **for** each training sample in a mini-batch **do**
13:     Calculate partial derivative with respect to the output: $\frac{\partial J}{\partial \phi_n}$.
14:     Run backward propagation to obtain the gradient with respect to the network parameters: $\Delta \Theta_n$.
15:     Accumulate gradients: $\Delta \Theta := \Delta \Theta + \Delta \Theta_n$.
16:   **end for**
17:   $\Theta^t := \Theta^{t-1} - \alpha_t \Delta \Theta$.
18: **until** converge.

---

We can assess the augmentation capability of DropRegion as follows. Assume a character image is divided into $L \times L$ regions by an elastic meshing technique, and $n$ ($n < L \times L$) regions are randomly disrupted. The number of possible combinations of the remaining regions is then $C_{L \times L}^n$. We add all possible combinations to yield no more than $n$ regions of disruption. We obtain a total of $\sum_{i=1}^n C_{L \times L}^i = 2^n$ variant samples. That is, the number of diverse samples is increased $2^n$ fold during the training phase, which demonstrates considerable augmentation capacity. As described previously, DropRegion disrupts the original topological structure of characters; nevertheless, it preserves the font information for consideration. Intuitively, using these partially destroyed samples, the model is guided to capture more invariant representation, regardless of the presence or absence of one partial region.

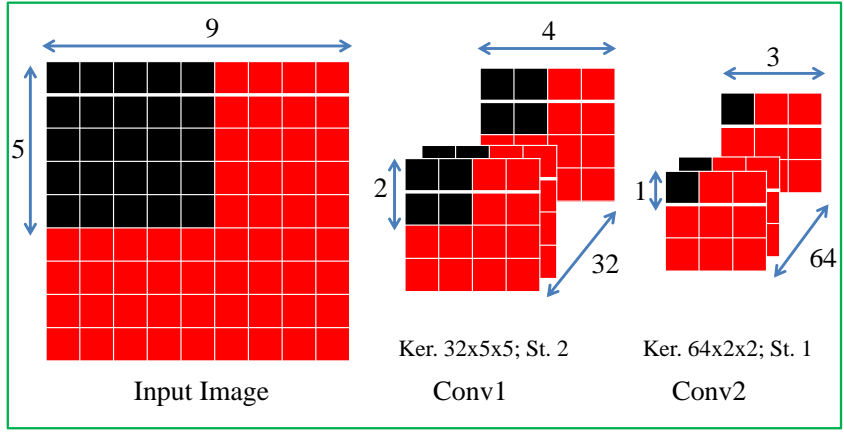

Fig. 4 Method by which DropRegion affects the convolutional layers. "Ker." represents the number of convolutional filters and their receptive field sizes ("num $\times$ size $\times$ size"), "St." is the stride size, and black pixels are zeroes

## 4 Single-character-based font recognition

We observed that more than one font is applied in a text block in many Chinese documents. To highlight a particular document author's intention, certain characters in a sentence may appear in a different font. For example, specific characters may be printed in boldface, while others in the same sentence may appear in a regular typeface. To handle these situations and provide greater flexibility, it is necessary to design a font recognizer for single unknown Chinese characters.

The parameters of the IFN designed for single-character font recognition are listed in Table 1. The network is abbreviated as *SingleChar-IFN*. This network contains four convolutional layers, six CCCP layers, one modified inception module, and one global average pooling layer. The settings for each of the convolutional layers ("conv") are given in three sub-rows. The first represents the number of convolutional filters and their receptive field sizes ("num $\times$ size $\times$ size"), the second indicates the convolution stride ("st"), and the third denotes spatial padding ("pad"). When stride is equal to 1 or no padding is executed in the convolution operation, the corresponding "st" or "pad" sub-row is omitted. The same is true for the CCCP layer. The settings for the maximum pooling are specified as their kernel size ("size $\times$ size") and stride size ("st"). A similar arrangement is used for the global average pooling layer, except that no stride is necessary, as the pooling is globally executed over the entire feature map. The setting for the dropout is specified as the dropout rate, which is in fact a regularization operation executed over the adjacent lower layer. *SingleChar-IFN* takes $64 \times 64$ gray images as input and outputs 25 dimensional vectors. The parameter values for DropRegion in *SingleChar-IFN* were set to $L = 5$ and $\gamma = 0.5$.

## 5 Text block-based font recognition

The basic idea of text block-based font recognition is initial segmentation of the text block into separate characters and recognition of the single character font. To this end, a CFR system based on character segmentation is described. The original image is transformed into a binary image by an adaptive thresholding algorithm. Next, dilation and erosion operations are performed on the binary images to remove noise and connect the different parts of one character. In addition, horizontal and vertical projection profiles are successively applied. The goal of using the horizontal projection profile is to determine each text line, while that of using the vertical projection profile is to locate characters. Once the characters are located, font classification is performed on each character through *SingleChar-IFN*. The final result is obtained by averaging 30 confidence vectors. Multiple samples are used together to make the decision. This can be understood as an ensemble method that improves the model robustness. However, segmentation-based text block font recognition is limited in that it depends on the segmentation procedure and thus increases the complexity of recognition methods. To address this issue, we developed a segmentation-free font recognition system called *TextBlock-IFN*. A flowchart of the system is shown in Fig. 5. The parameters of *TextBlock-IFN* are outlined in Table 2. For the training phase, $128 \times 128$ patches are randomly cropped from one $320 \times 320$ text block image (a detailed description of the text block samples are given in Section 6). These $128 \times 128$ patches are then disrupted by the DropRegion method. In a manner similar to the process for training *SingleChar-IFN*, the DropRegion parameter values in *TextBlock-IFN* are set to $L = 5$ and $\gamma = 0.5$.

For the testing phase, we designed a multiple-decision fusion method combined with a sliding window approach. A sliding window was used to crop the image patches. The window size was $128 \times 128$, and the stride was 64 pixels. Using the sliding window, 20 cropped image patches were obtained from one text block. These cropped image patches were used as inputs to *TextBlock-IFN*. The output was a 25-dimensional (25-D) confidence vector for each cropped patch. An averaging ensemble was adopted to enhance the model robustness. We applied element-wise accumulation to the 25-D confidence vectors.

Table 1 Network parameters for single-character font recognition (*SingleChar-IFN*).

| SingleChar-IFN | | |
|---|---|---|
| Type | Settings | Output size |
| Input |  | $64 \times 64 \times 1$ |
| Conv 1 | $96 \times 7 \times 7$ | $58 \times 58 \times 96$ |
| CCCP 1_1 | $96 \times 1 \times 1$ | $58 \times 58 \times 96$ |
| CCCP 1_2 | $96 \times 1 \times 1$ | $58 \times 58 \times 96$ |
| Max - pooling | $3 \times 3$, st. 2 | $29 \times 29 \times 96$ |
| Conv 2 | $256 \times 7 \times 7$ | $23 \times 23 \times 256$ |
| CCCP 2_1 | $256 \times 1 \times 1$ | $23 \times 23 \times 256$ |
| CCCP 2_2 | $256 \times 1 \times 1$ | $23 \times 23 \times 256$ |
| max - pooling | $3 \times 3$, st. 2 | $11 \times 11 \times 256$ |
| **Modified Inception** | detailed in Fig.1 | $11 \times 11 \times 604$ |
| Conv 3 | $512 \times 3 \times 3$, pad 1 | $11 \times 11 \times 512$ |
| CCCP 3_1 | $512 \times 1 \times 1$ | $11 \times 11 \times 512$ |
| CCCP 3_2 | $512 \times 1 \times 1$ | $11 \times 11 \times 512$ |
| Dropout | 0.5 |  |
| Conv 4 | $25 \times 1 \times 1$ | $11 \times 11 \times 25$ |
| Global Ave Pooling | $11 \times 11$ | $1 \times 1 \times 25$ |

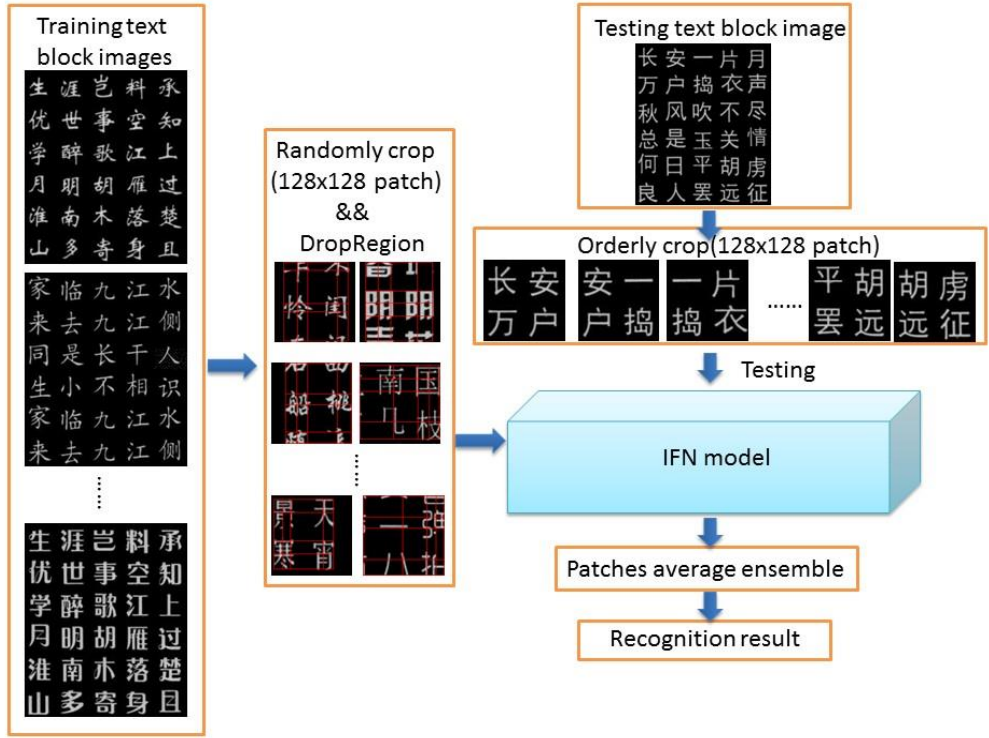

Fig. 5 Flowchart of segmentation-free text block font recognition.

## 6 Experiments

### 6.1 Dataset and implementation

To the best of our knowledge, no publicly available datasets exist for Chinese character font recognition. We therefore collected data samples from scanned documents to build a scanned Chinese font database, called SCF_DB. The database was constructed according to the following five steps. (1) A total of 3,866 commonly used Chinese characters were edited with 25 font styles using the Microsoft Word software. The 25 fonts were Hei, YaHei, XiHei, YueHei, MeiHei, YaYuan, XingKai, Kai, FangSong, Song, ZhongSong, ZongYi, HuoYi, CuQian, GuangBiao, HuangCao, HuPo, Ling Xin, Shu, WeiBei, XinWei, YaoTi, YouYuan, LiShu, and ShuangXian. (2) The characters were then printed on paper and scanned as images. (3) The Hough line detector algorithm was used to align the images (4)An image dilation method was used to connect adjacent characters. As a result, each text line was located using a horizontal projection profile, and each character was located using a vertical projection profile. (5) The character images, together with a number of annotations (including character category, font category, etc.), were saved in the database.

Table 2 Architectures of the text block network (*TextBlock-IFN*).

| TextBlock-IFN | | |
|---|---|---|
| **Type** | **Settings** | **Output size** |
| Input | | $128 \times 128 \times 1$ |
| Conv 1 | $96 \times 7 \times 7$, st. 2 | $61 \times 61 \times 96$ |
| CCCP 1_1 | $96 \times 1 \times 1$ | $61 \times 61 \times 96$ |
| CCCP 1_2 | $96 \times 1 \times 1$ | $61 \times 61 \times 96$ |
| Max - pooling | $3 \times 3$, st. 2 | $30 \times 30 \times 96$ |
| Conv 2 | $256 \times 7 \times 7$ | $24 \times 24 \times 256$ |
| CCCP 2_1 | $256 \times 1 \times 1$ | $24 \times 24 \times 256$ |
| CCCP 2_2 | $256 \times 1 \times 1$ | $24 \times 24 \times 256$ |
| max - pooling | $3 \times 3$, st. 2 | $11 \times 11 \times 256$ |
| **Modified Inception** | detailed in Fig. 1 | $11 \times 11 \times 604$ |
| Conv 3 | $512 \times 3 \times 3$ | $11 \times 11 \times 512$ |
| CCCP 3_1 | $512 \times 1 \times 1$ | $11 \times 11 \times 512$ |
| CCCP 3_2 | $512 \times 1 \times 1$ | $11 \times 11 \times 512$ |
| Dropout | 0.5 | |
| Conv 4 | $25 \times 1 \times 1$ | $11 \times 11 \times 25$ |
| Global Ave Pooling | $11 \times 11$ | $1 \times 1 \times 25$ |

We evaluated the proposed method using cases of single characters and text blocks. The corresponding settings for these two types of experiments were the following. For single-character-based font recognition, top *TrNum* samples were chosen from each font to build the training dataset, and the remaining samples were used as the testing dataset. Each single character image was resized to $60 \times 60$. Then, each image was padded with two pixels in the boundary, so that the input image for *SingleChar-IFN* was $64 \times 64$.

For text block-based font recognition, 320 text blocks were composed for each font using 320 Tang poems. That is, one Tang poem corresponded to one text block consisting of six lines. Each line contained five characters. The characters not belonging to the set of 3,866 characters were ignored. If the character number of one poem was less than 30, it was repeated from the beginning of the poem. Figure 6 shows text block samples.

We implemented our algorithm based on the Caffe [42] framework. The learning rate was updated as follows:

$$lr = base\_lr \times \left(1 - \frac{iter}{max\_iter}\right)^{factor}, (8).$$

where $base\_lr$, $max\_iter$, and factor were set to 0.01, 1.5e5, and 0.5, respectively. The momentum was set to 0.9, and the weight decay rate was set to 2e-4.

### 6.2 Experimental results

#### 6.2.1 Evaluation of the effect of IFN network structure

We evaluated the effectiveness of the proposed IFN network structure in contributing to the overall performance achieved in Chinese character font recognition. Four baseline network architectures, summarized as shown in Fig. 7, were used for comparison. The four are denoted Basic-CNN, DeepFontNoD, MDLSTM [44] and GoogleNet [16]. Basic-CNN is a standard CNN architecture consisting of four stacked convolutional layers, optionally followed by max-pooling, and then, ending with a full connection layer feeding into a final softmax layer. The architectural parameters of Basic-CNN are the same as that of IFN at the corresponding layer. DeepFontNoD

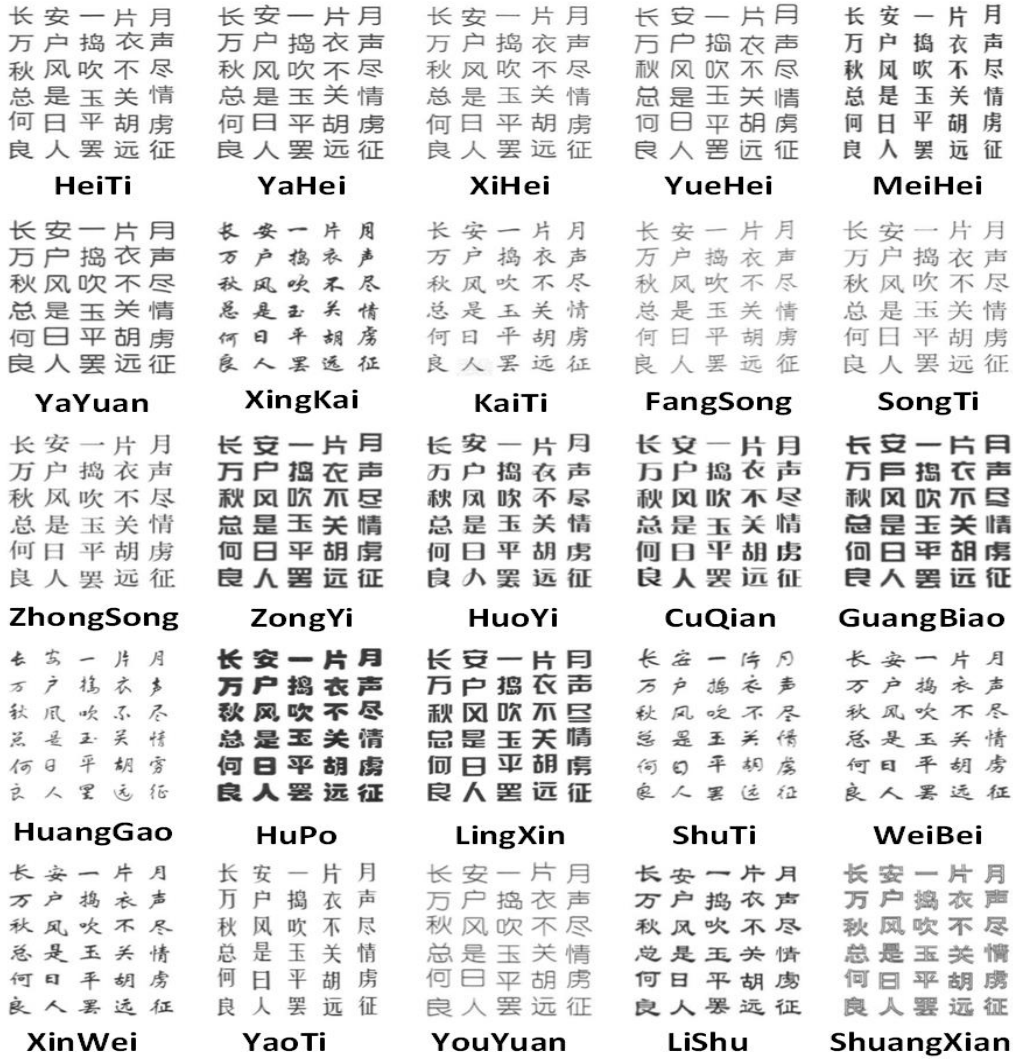

Fig. 6 Text block samples of 25 fonts.

has the same basic network structure as the DeepFont system [1], with low-level unsupervised and high-level supervised sub-network decomposition removed. This structure is similar to that of AlexNet [13], which serves as an example for the adoption of a conventional CNN structure instance in the field of font recognition. For MDLSTM, we adopted the same network hierarchy by repeatedly composing MDLSTM layers with feed-forward layers, as in [45]. GoogleNet has almost the same settings as in [16], except that the input size is different (64 ×64 vs. 224 ×224). The hyper-parameters for the last average pooling layer were adjusted from "7 × 7, st. 1" to "2 × 2, st. 1," as shown in Fig. 7. The dashed rectangle in the figure represents a repetitive network building block, which can be removed from the corresponding network to obtain a simplifier model for adjusting to training sample scale. For example, when $TrNum$ = 200 samples were chosen from each font to build a training set, only one NIN building block was retained in the subsequent experiment setting to adapt to the case of a small sample size. Two NINs were retrain

-ed in the case of $TrNum$ = 400, and three were retained in the case of $TrNum$=1000. Besides, a similar adjustment to the building block size was also adopted for Basic-CNN when comparison experiments were conducted for the purpose of verification of the network structure design.

To ensure a fair comparison, we used Dropout for all the models, rather than DropRegion. The experimental results for the task of single-character-based font recognition are summarized in Table 3. The results show that the IFN architecture almost always outperformed all four of the baseline networks for each font for sample sizes of 200, 400, and 1,000. The results confirm the effectiveness of our IFN design. As Table 3 shows, Basic-CNN performs much better than DeepFontNoD, even they are homogenous. This may be because Basic-CNN naïvely adjusts the model size by adding or removing building blocks to or from the network structure to adapt to the sample scale. The same model size adjustment scheme is adopted in both IFN and Basic-CNN, but IFN results in a clear improvement in the accuracy in comparison to Basic-CNN. Therefore, it is of critical importance that the architectural choice is elegantly designed for the specific font recognition task. Compared to MDLSTM, the relative increase in font category recognition accuracy achieved by IFN is up to 22% when $TrNum$ = 200. The accuracy of GoogleNet was greater than our IFN by approximately 1.39% when $TrNum$ = 1000. However, the IFN network has many fewer parameters than GoogleNet (20 M vs. 34 M). To summarize the comparative evaluation of the five various architectures, the design of our IFN architectural elements

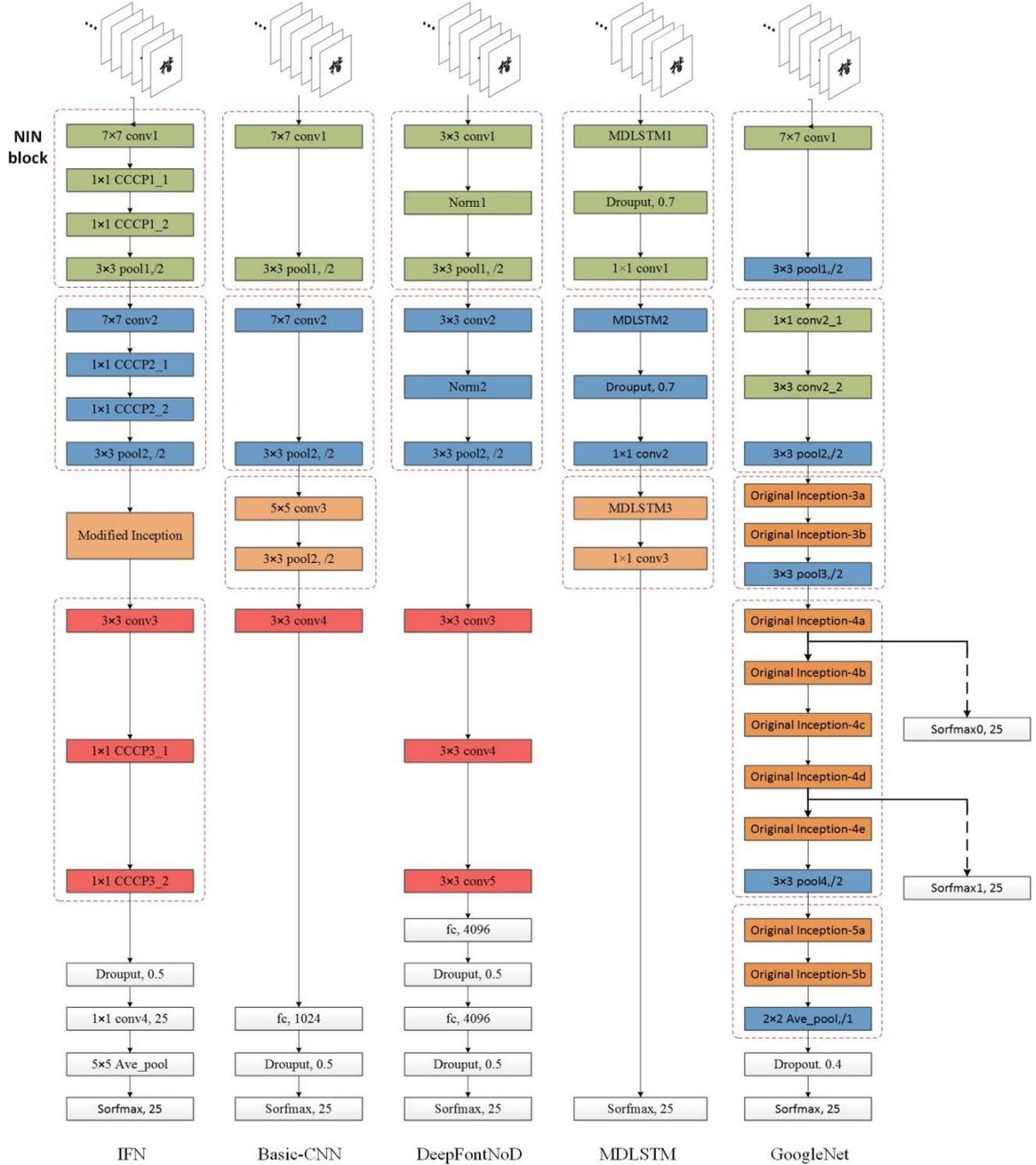

Fig. 7 Five network architectures used in our experiments. This plot shows the network structure for the five different baseline models

### 6.2.2 Evaluation of DropRegion in comparison to Dropout

We evaluated the regularization effect of Dropout and DropRegion using *SingleChar-IFN*. In the experi-ments, each single image was divided into 5 × 5 regions, and a maximum of 13 regions were randomly dropped. The results are shown in Table 4. It can be seen that Dropout improved the accuracy by 4.18%, 1.36%, and 0.01% when 200, 400, and 1,000 training samples, respectively, were used for each font. DropRegion boosted the accuracy by 12.10%, 7.74%, and 5.27%, respectively, for the same numbers of training samples, exhibiting better regularization performance than Dropout. We also investigated the performance of a combination of DropRegion and Dropout. As Table 4 shows, the combination further improved the classification accuracy. This suggests that Dropout and DropRegion are complementary when used together. It is worth mentioning that the recognition accuracies we achieved exceeded that achieved with Tao's method [43], even when we used training sets only half the size of those used with Tao's method (e.g., 2,000 for each font), based on total data sets of almost the same size.

Table 3 Recognition accuracies (%) of IFN and other baseline networks (Basic-CNN, DeepFontNoD, MDLSTM, and GoogleNet) for single-character images. *TrNum* is the number of training samples for each font.

| TrNum | IFN | Basic-CNN | DeepFontNoD | MDLSTM | GoogleNet |
|---|---|---|---|---|---|
| 200 | **89.34** | 83.64 | 73.00 | 73.52 | 80.57 |
| 400 | **91.54** | 88.06 | 80.51 | 78.20 | 86.05 |
| 1,000 | 93.59 | 92.85 | 90.00 | 81.28 | **94.98** |

Table 4 Font recognition accuracy (%) of DropRegion and Dropout on single characters

| TrNum | None | Dropout | DropRegion | DropRegion+Dropout |
|---|---|---|---|---|
| 200 | 85.16 | 89.34 | 97.26 | **97.31** |
| 400 | 90.18 | 91.54 | 97.92 | **98.63** |
| 1000 | 93.59 | 93.60 | 98.86 | **98.98** |

Table 5 Recognition accuracies (%) of different methods for single character images. *TrNum* is the number of training samples for each font.

| TrNum | SingleChar-IFN | SingleChar-IFN+ DropRegion (fixed mesh) | SingleChar-IFN+ DropRegion (elastic mesh) |
|---|---|---|---|
| 200 | 89.34 | 97.20 | **97.31** |
| 400 | 91.54 | 98.30 | **98.63** |
| 1,000 | 93.59 | 98.96 | **98.98** |

Table 6 Evaluation of recognition accuracy (%) achieved by varying the number of dropped regions.

| n | 1 | 2 | 3 | 4 | 5 | 6 | 7 | 8 |
|---|---|---|---|---|---|---|---|---|
| Accuracy | 90.16 | 94.46 | 95.51 | 95.80 | 95.98 | 96.19 | 96.28 | 96.73 |
| n | 9 | 10 | 11 | 12 | 13 | 14 | 15 | 16 |
| Accuracy | 96.89 | 96.97 | 97.12 | 97.20 | **97.31** | 96.92 | 96.80 | 96.67 |
| n | 17 | 18 | 19 | 20 | 21 | 22 | 23 | 24 |
| Accuracy | 96.59 | 96.35 | 96.09 | 95.94 | 95.87 | 94.82 | 93.84 | 93.49 |

Table 7 Recognition accuracy (%) of multiple groups of segmentation-free text block font recognition experiments. G1 to G8 denote eight groups of experiments. *TrNum* is the number of training samples for one font. *TsNum* is the number of testing samples for one font.

| Dataset | G1 | G2 | G3 | G4 | G5 | G6 | G7 | G8 | Average |
|---|---|---|---|---|---|---|---|---|---|
| TrNum = 30  TsNum = 10 | 99.6 | 99.92 | 99.9 | 99.9 | 99.96 | 99.58 | 99.66 | 99.72 | 99.78 |
| TrNum = 20  TsNum = 20 | 99.57 | 99.4 | 99.29 | 98.93 | 99.82 | 99.6 | 98.3 | 99.3 | 99.28 |

Table 8 Recognition accuracy (%) of the proposed method and other state-of-the-art methods.

| Methods | TrNum = 30 TsNum = 10 | TrNum = 20 TsNum = 20 |
|---|---|---|
| Proposed (segmentation-free) | **99.78** | **99.28** |
| Proposed (segmentation) | 96.90 | 96.80 |
| LBP+SVM | 95.35 | 94.20 |
| Gabor+WED[3] | 95.45 | 94.48 |
| SDIP[5] | 93.00 | 91.60 |
| MFA[34] | 91.60 | 90.10 |

### 6.2.3 Evaluation of fixed and elastic mesh techniques

We evaluated three different methods using training samples of different sizes. The methods were *SingleChar-IFN*, *SingleChar-IFN*+DropRegion (fixed mesh), and *SingleChar-IFN*+DropRegion (elastic mesh). The value of the parameter $L$ was fixed or the elastic meshing division was set to 5, and the value of the parameter $\gamma$ was set to 0.5.

The experimental results are presented in Table 5. For $TrNum$ = 200, 400, and 1,000, the DropRegion method with a fixed mesh improved the recognition accuracy by 7.86%, 6.76%, and 5.37%, respectively. The DropRegion method with an elastic mesh improved the recognition accuracy by 7.97%, 7.09%, and 5.39%, respectively. These results suggest that, regardless of whether the fixed or elastic meshing method is used, DropRegion improves the capacity of representation learning, especially when the training samples are small. Furthermore, the fixed and elastic meshing techniques boost the recognition accuracy to varying degrees (7.86% vs. 7.97%, 6.76% vs. 7.09%, and 5.37% vs. 5.39%, respectively). These two meshing techniques are both effective, but the results suggest that the elastic meshing method is slightly superior to the fixed meshing method.

### 6.2.4 Investigation of number of dropped regions

We also investigated the effect of the number of dropped regions. $TrNum$ is set as 200. Each single character image is divided into $5 \times 5$ regions. The maximum number of dropped regions, denoted as $n$, is varied from 1 to 24. Different $n$ represent different increased folds of the variant sample. The experimental results are shown in Table 6. The best recognition accuracy (97.31%) is achieved when the number of dropped regions is 13. When the number of dropped regions is less or more than 13 (between 1 and 24), the recognition accuracy is much lower than 97.31%. This is because few stochastic variations are introduced when the number of dropped regions is decreased, and the font information is not sufficient for learning a robust model when the number of dropped regions increases.

### 6.2.5 Segmentation-free-based text block font recognition

We performed eight sets of experiments to evaluate segmentation-free text block font recognition. We followed the dataset partition scheme introduced by Tao et al. [43] The results are shown in Table 7. An average accuracy of 99.78% was achieved when *TrNum* and *TsNum* were 30 and 10, respectively, and an average accuracy of 99.28% was achieved when *TrNum* and *TsNum* were 20 and 20, respectively.

We compared our segmentation-free recognition system with the segmentation-based system described in section 5 and four representative state-of-the-art systems, LBP+SVM, Gabor+WED, [3] sparse discriminative information preservation [5] (SDIP), and marginal Fisher's analysis [34] (MFA). As Table 8 shows, the proposed segmentation-free-based method achieved the highest accuracy of 99.78% when the training number (*TrNum*) was 30 and the test number (*TsNum*) was 10. Similarly, the highest accuracy of 99.28% was achieved when *TrNum* and *TsNum* were both 20. These results can be attributed to the design of the deep IFN and the design of DropRegion. The proposed segmentation-based method achieved 96.90% and 96.80% accuracy for two different training samples, thereby ranking second among the systems. Compared with other handcrafted features (i.e., LBP and Gabor features), deep IFN automatically discovers the features of font recognition within an end-to-end learning framework, which proves to be a powerful method for feature learning. The proposed DropRegion method improves the performance of IFN because it significantly increases the number of training samples and can be regarded as an effective regularization technique.

## 7 Conclusions

In this paper, we have presented a new method, called DropRegion-IFN, for Chinese font recognition. This method highlights the elegant design of a deep font network, namely IFN, which integrates a modified inception module, CCCP layers, and global average pooling. DropRegion, a new data augmentation and regularization technique, is proposed for seamless embedment in an IFN framework, enabling an end-to-end training approach and enhancing model generalization. The results of experiments conducted using the scanned Chinese font database SCU-DB show that a character recognition accuracy of 97.31% could be achieved using relatively few training examples (5,000 training samples from among a total of 96,650 samples). When 25,000 training samples were used, an accuracy of 98.98% was achieved, setting a new benchmark for accuracy in Chinese character font recognition. A segmentation-free text block-based Chinese font recognition system using DropRegion-IFN outperformed all of the methods to which it was compared, suggesting that the combination of the IFN and DropRegion training is a good solution for Chinese font recognition.

In future research, we will construct a more difficult Chinese font dataset by collecting labeled real-world text images that cover a larger set of Chinese font classes and more diverse samples, with many challenging factors such as perspective distortions, cluttered backgrounds, very low resolution, and considerable noise. We would like to extend the DropRegion-IFN method to handle the practical cases that such a dataset would provide, even if the font class samples are scarce. In addition, DropRegion will be developed to gear it toward being a scalable data-driven machine learning approach suitable for use in solving other computer vision problems, such as object detection and scene text recognition in the case of sample scarcity.